**EyeFound: A Multimodal Generalist Foundation Model for Ophthalmic Imaging**

Danli Shi, MD[1,2#], Weiyi Zhang, MS[1#], Xiaolan Chen, MD[1], Yexin Liu, MS[1], Jiancheng Yang, PhD[3], Siyu Huang, PhD[4], Yih Chung Tham, MD, PhD[5], Yingfeng Zheng, MD, PhD[6], Mingguang He, MD[1,2,7*]

**Author Affiliations:**

1. School of Optometry, The Hong Kong Polytechnic University, Kowloon, Hong Kong.

2. Research Centre for SHARP Vision (RCSV), The Hong Kong Polytechnic University, Kowloon, Hong Kong.

3. Swiss Federal Institute of Technology Lausanne (EPFL), Lausanne, Switzerland.

4. Clemson University, USA.

5. Yong Loo Lin School of Medicine, National University of Singapore; Singapore Eye Research Institute, Singapore.

6. State Key Laboratory of Ophthalmology, Zhongshan Ophthalmic Center, Sun Yat-sen University, Guangdong Provincial Key Laboratory of Ophthalmology and Visual Science, Guangdong Provincial Clinical Research Center for Ocular Diseases, Guangzhou 510060, China.

7. Centre for Eye and Vision Research (CEVR), 17W Hong Kong Science Park, Hong Kong.

[#] Contributed equally

**Correspondence**

**\*Prof. Mingguang He,** MD, PhD., Chair Professor of Experimental Ophthalmology, School of Optometry, The Hong Kong Polytechnic University, Kowloon, Hong Kong SAR, China. Email: mingguang.he@polyu.edu.hk

**\*Dr. Danli Shi,** MD, PhD., School of Optometry, The Hong Kong Polytechnic University, Kowloon, Hong Kong SAR, China. Email: danli.shi@polyu.edu.hk.

**Abstract**

Artificial intelligence (AI) is vital in ophthalmology, tackling tasks like diagnosis, classification, and visual question answering (VQA). However, existing AI models in this domain often require extensive annotation and are task-specific, limiting their clinical utility. While recent developments have brought about foundation models for ophthalmology, they are limited by the need to train separate weights for each imaging modality, preventing a comprehensive representation of multi-modal features. This highlights the need for versatile foundation models capable of handling various tasks and modalities in ophthalmology. To address this gap, we present EyeFound, a multimodal foundation model for ophthalmic images. Unlike existing models, EyeFound learns generalizable representations from unlabeled multimodal retinal images, enabling efficient model adaptation across multiple applications. Trained on 2.78 million images from 227 hospitals across 11 ophthalmic modalities, EyeFound facilitates generalist representations and diverse multimodal downstream tasks, even for detecting challenging rare diseases. It outperforms previous work RETFound in diagnosing eye diseases, predicting systemic disease incidents, and zero-shot multimodal VQA. EyeFound provides a generalizable solution to improve model performance and lessen the annotation burden on experts, facilitating widespread clinical AI applications for retinal imaging.

**Introduction**

Medical artificial intelligence (AI) has made significant strides in recent years, particularly with the evolution of deep learning techniques.[1] However, most medical AI models still rely heavily on specific-modality annotated data and are designed for specific tasks.[2,3] Such models require substantial annotation effort and often result in a scarcity of data for certain classes, leaving a large amount of unlabeled data unused and wasted. To address these issues, foundation models have been proposed to perform more generalized AI tasks.[4,5] These models are pretrained on vast amounts of data to extract generalizable feature representations, which can be easily adapted to specific tasks by fine-tuning on a small amount of data with explicit labels. This approach is becoming a new trend in modern medical AI.[6]

In ophthalmology, a notable advancement in foundational models is RETFound.[5] This model uses self-supervised learning, specifically the generative pretraining method known as masked autoencoder (MAE) [7], to learn retinal feature-specific representations from millions of unlabeled real-world data. MAE works by attempting to reconstruct original retinal images from highly masked image patches with limited visible information. This forces the model to learn the data distribution related to critical retinal structures like vessels and the optic nerve, which are essential for diagnosing retinal diseases. RETFound has shown promising results in various downstream tasks, including retinal and systemic diseases, and has improved training efficiency. However, RETFound was trained separately for color fundus photographs (CFP) and optical coherence tomography (OCT), and its downstream tasks are tested solely on classification. In practical ophthalmology, a wide range of image modalities are used, each based on different technologies such as light, electronics, lasers, ultrasound, etc., resulting in significant differences between the images. Foundation models limited to narrow modalities may restrict their applicability in broader clinical settings. Furthermore, the features from different modalities were not jointly learned, which means it misses out on the potential benefits of shared cross-modal information that could improve feature representation and downstream accuracy and efficiency. [8-10]

In this paper, we propose EyeFound, a generalist foundational model for multimodal and multi-task purposes in ophthalmology. We have constructed a large-scale ophthalmology multimodality dataset comprising 11 commonly used clinical image modalities, including CFP, fundus fluorescein angiography (FFA), indocyanine green angiography (ICGA), fundus autofluorescence (FAF), RetCam, Ocular ultrasound, OCT, slit-lamp, external eye photo, specular microscope, and corneal topography. This dataset covers a wide range of ophthalmic diseases, ages, and camera types collected from 227 hospitals across China. We used this dataset to develop EyeFound and tested it on multiple downstream datasets, including ocular disease classification, systemic disease prediction, medical report generation, and visual question answering (VQA).

**Methods**

EyeFound uses an extensive dataset for pretraining, enabling it to learn rich representations of ophthalmic images and capture a broad spectrum of anatomical structures and pathologies across 11 modalities. It is then applied to various datasets for different downstream tasks. Figure 1 shows the study diagram.

**Data for pretraining EyeFound**

We collected a large amount of unannotated ophthalmic images across 227 hospitals in China, totaling 3,281,281 images. We excluded low-quality images in CFP, FFA, and ICGA by extracting vessels, [11] where images with detachable vessel ratios less than 0.04 (CFP), and less than 0.01 (FFA, ICGA) were excluded. We manually checked images from other modalities, they were diverse and were less likely to be of quality because they were captured in the clinical setting and will be of optimal quality to be distributed to patients, therefore we did not use a specific method for those images.

**Data for downstream tasks**

**Retinal diseases classification**

For comparison with RETFound, we used the same eight publicly available ophthalmic datasets and another two public datasets. There were nine datasets for CFP, including diabetic retinopathy (DR), glaucoma, and multi-class eye disease diagnosis. Among them, Kaggle APTOS-2019 (India), IDRID (India) [12], and MESSIDOR-2 (France) were used for DR diagnosis, the labels were based on the International Clinical Diabetic Retinopathy Severity Scale, indicating five stages from no DR to proliferative DR. PAPILA (Spain) [13] and Glaucoma Fundus (South Korea)[14] were used for glaucoma, including three classification labels: non-glaucoma, early glaucoma (suspected glaucoma), and advanced glaucoma. The multi-class eye disease CFP datasets included the JSIEC (China)[15], Retina dataset, and (Multi-Label Retinal Diseases) MuReD dataset[16]. JSIEC consists of 1,000 images, including 39 common fundus diseases and conditions. Retina has labels for normal, glaucoma, cataract, and retinal diseases. MuReD was collected from three different sources with 20 different labels. For OCT, we used two datasets, the OCTID (India) [17] and the OCTDL[18] dataset for multi-class disease diagnosis. OCTID includes 572 OCT scans labeled as normal, macular hole, age-related macular degeneration (AMD), central serous chorioretinopathy (CSCR), and DR. OCTDL includes labels for AMD, Diabetic Macular Edema (DME), Epiretinal Membrane (ERM), normal, Retinal Artery Occlusion (RAO), Retinal Vein Occlusion (RVO), Vitreomacular Interface Disease (VID). Supplementary Table 1 presents the characteristics of the datasets.

**Multimodal multilabel ophthalmic disease classification**

The AngioReport dataset (APTOS2023, Thailand)[19] contains FFA and ICGA images collected from routine retinal clinics, The images are annotated with retinal disease information and detailed findings of fluorescin manifestation, and they could be used for report generation or multi-label diagnostic performance evaluation. This dataset consists of over 50,000 angiographic images, for efficiency, we used the test subset for performing experiments. The dataset includes 142 retinal conditions, such as DR, choroidal neovascularization (CNV), and RVO.

Retina Image Bank[20], a project by the American Society of Retina Specialists, is a vast open-access library of more than 25,000 unique and downloadable retina images uploaded by ophthalmologists. Here we scraped the images and findings from the web page and built a custom dictionary to map different expressions of diseases into a standard one using keyword-matching with regular expressions. The mapped labels contain hierarchy, for example for mild DR, the label is "DR, mild DR". We excluded images that are not routine retinal examinations, for example, schematic cartoon draws, histology, and pathology images. For efficiency, we included images uploaded from 2019-2023. Additionally, we excluded conditions that are less than 50 occurrences. Supplementary Table 2 presents the characteristics of the datasets. Supplementary Figure 1 shows the distribution of retinal conditions of the Retina Image Bank.

**Systemic disease prediction**

The systemic disease prediction task focuses on a 5-year onset prediction of cardiovascular and neurological diseases using the UK Biobank[21] dataset. The UK Biobank consists of 502,665 UK residents aged between 40 and 69, who are registered with the National Health Service. Among all participants, 82,885 underwent CFP examination, resulting in a collection of 171,500 retinal images. We used the algorithm-defined outcomes (Category 42) for predicting four major systemic diseases: stroke, dementia, Parkinson's disease (PD), and myocardial infarction (MI), based on CFP. For each patient, we included only the retinal images of the right eye during one visit to avoid potential biases caused by inconsistent individual visits. Supplementary Table 3 presents the characteristics of the datasets.

**Visual Question Answering**

OphthalVQA[22] is a dataset of ocular multimodal images from China. This dataset included slit-lamp, scanning laser ophthalmoscopy (SLO), CFP, OCT, FFA, and ocular ultrasound images. For each modality, ten images representing distinct diagnoses were carefully selected to form the VQA dataset, resulting in a total of 60 images and 600 free-form QA pairs generated by ophthalmologists. These images exhibited typical disease manifestations used by ophthalmologists for diagnosis. Supplementary Table 2 presents the characteristics of the dataset.

For all downstream datasets, we divided the images into training, validation, and testing sets at a ratio of 55:15:30%.

**EyeFound Model Architecture and Implementation**

Data processing: We cropped out the extra black background in the images using a custom threshold method, where the pixels below the threshold will be set to zeros, and the images will be cropped based on the ranges of non-zeros pixels, the threshold for CFP was 15, and the threshold for OCT was 30. The cropped images were resized to a resolution of 256 × 256 via cubic interpolation.

Data augmentation: In the model training phase, we adhered to the data augmentation techniques analogous to those used in masked autoencoders, which include random cropping (ranging from a minimum of 20% to the entirety of the image) and resizing the cropped patches to 224 × 224 pixels, along with random horizontal flips and image normalization.

Architecture: We use the masked autoencoder as our framework, and used the weight from RETFound as initialization. Specifically, it consists of an encoder and a decoder. The encoder utilizes a large vision transformer (ViT-large), which consists of 24 Transformer blocks and has an embedding vector size of 1,024. The decoder utilizes a smaller vision transformer (ViT-small), which consists of 8 Transformer blocks and has an embedding vector size of 512. The encoder takes unmasked patches (patch size of 16 × 16) as input and projects it into a feature vector of size 1,024. The 24 Transformer blocks, including multi-head self-attention and multi-layer perceptrons, take the feature vector as input and generate high-level features. The decoder inserts masked virtual blocks into the extracted high-level features as model input and then reconstructs image patches after a linear projection. During the training process, the objective is to reconstruct retinal images from highly masked versions. The mask ratio was 0.80. The batch size is 64. The total training epoch is 50 epochs and the first 15 epochs are for learning rate warm-up (learning rate ranging from 0 to $1 \times 10^{-3}$). We selected the best-performing epoch through downstream performance on the validation set and conducted 5 trials with 5 different random seeds on each downstream dataset.

**Adaptation to classification downstream tasks**

For classification tasks, we rely solely on the encoder (ViT-large) of the foundation model, omitting the decoder. This encoder extracts key features from images, which are then fed into a multilayer perceptron (MLP). The MLP computes the probability distribution across disease categories, and the number of categories determines the neurons in the final layer of the MLP. To prevent overfitting, we integrate label smoothing, aligning the output distribution with the categorical labels.

Specifically, for single-label classification tasks, we use a batch size of 16 and train for a total of 50 epochs. The first ten epochs implement a learning rate warm-up from 0 to $5 \times 10^{-4}$, followed by a cosine annealing schedule reducing the learning rate from $5 \times 10^{-4}$ to $1 \times 10^{-6}$ over the remaining 40 epochs. For multi-label classification tasks in AngioReport and Retina Image Bank, we use a batch size of 4, train for 30 epochs, and set the learning rate to 0.01. After each epoch, we evaluate the model on the validation set, saving the model weights with the highest AUROC for both internal and external assessments.

**Adaptation to visual question-answering downstream task**

We utilize the ViT-large encoder from the foundation model to extract image features, which are then concatenated with text features. The combined feature is fed into the language model Vicuna (Llama 2) [23] for language generation, performing VQA. For fine-tuning, we leverage a subset of the Retina Image Bank (2019-2023), comprising 1,215 samples. Subsequently, we assess the fine-tuned model on the OphthalVQA dataset for zero-shot VQA testing. Training is conducted in PyTorch on a dual V100 setup, with the fine-tuning process employing Low-Rank Adaptation (LoRA) [24] and a batch size of 8. Training spans three epochs with an initial learning rate of 2e-5, and we incorporate cosine annealing for dynamic learning rate adjustments. Evaluation is carried out directly on the final epoch.

**Explanations for EyeFound**

Using the encoder from EyeFound, we performed masking and reconstruction on images from various modalities to assess whether the model learned different features within retinal images across different modalities.

**Computational resources**

During the pretraining phase of EyeFound, we employed two NVIDIA Tesla V100 (32GB) GPUs, with the entire training process taking approximately one week to complete. For the fine-tuning of EyeFound for downstream tasks, we utilized a single NVIDIA Tesla V100 GPU.

**Evaluation and statistical analysis**

For classification tasks, we evaluated performance using the AUROC and AUPR metrics, which measure the classification performance from the receiver operating characteristics and precision-recall curves, respectively. For binary classification tasks, such as diagnosing ocular diseases, AUROC and AUPR were calculated in a binary setting. For multi-class classification tasks, such as five-stage DR and multi-class disease diagnosis, we calculated the AUROC and AUPR for each disease class and averaged (macro) them to obtain the general AUROC and AUPR.

For VQA tasks, classification-based metrics including exact match score, F1 score, precision and recall, and language-based metric Bilingual Evaluation Understudy (BLEU)[25] were used.

For each task, we conduct 5 random splits and train the model with varying random seeds to mitigate the effects of randomness. We calculate the mean of the performance over the five iterations and calculate the standard error to obtain the 95% confidence interval (CI) using 1.96 × standard error. We use the two-sided t-tests between the performance of EyeFound and RETFound AUROC to show whether significant differences exist.

**Ethics Statement**

This retrospective study used deidentified ophthalmic images and public datasets, therefore informed consent is not necessary. The study adheres to the tenets of the Declaration of Helsinki and was approved by the institutional review board of the Hong Kong Polytechnic University (HSEARS20240202004).

**Results**

**Development of EyeFound**

After quality control, we included 2,777,593 images, including 1,168,922 (42.1%) CFP, 768,622 (27.7%) FFA images, 287,155 (10.3%) slitlamp images, 264,934 (9.5%) ICGA images, 149,352 (5.4%) ocular ultrasound images, 54,126 (1.9%) OCT images, and 84,482 (3.0%) other modalities. Table 1 shows the dataset characteristics.

**Performance of Ophthalmic Disease Diagnosis with CFP and OCT**

We used public datasets to compare the diagnostic performance of EyeFound with RETFound for various eye diseases in both CFP and OCT imaging modalities. EyeFound performs comparably with RETFound for most datasets (P>0.05), occasionally outperforming it, such as in the case of Glaucoma fundus with an AUROC of 0.955 (95% CI 0.941, 0.969) vs 0.943 (95% CI 0.941, 0.945) for RETFound. All quantitative results are provided in Table 2.

For external evaluation, we evaluated the performance of EyeFound on DR datasets (MESSIDOR-2, IDRID[12], and APTOS-2019). We conducted cross-evaluations among the three datasets by fine-tuning models on one dataset and evaluating their performance on the others. EyeFound achieved a better performance in some cross-evaluations when compared to RETFound (Supplementary Table 4). For instance, when fine-tuned on the APTOS-2019 dataset, EyeFound achieved an AUROC of 0.803 and 0.769, along with AUPR of 0.442 and 0.433 on the IDRID and MESSIDOR-2 datasets, respectively. These performance metrics were higher than those achieved by RETFound.

**Performance of Multi-label Ophthalmic Disease Classification with angiographic images**

We conducted multi-label disease classification on angiographic ophthalmic images of FFA and ICGA using the AngioReport (APTOS2023)[19] dataset (test subset) and compared the results with RETFound, as in Table 2. EyeFound effectively distinguished 142 different retinal conditions, with AUROC=0.707 and AUPR=0.297, significantly outperforming RETFound with AUROC=0.693 and AUPR=0.295 (P<0.001).

**Performance of Multi-label Ophthalmic Disease Classification with Retina Image Bank**

Considering EyeFound was developed to accommodate a wide range of imaging conditions and ophthalmic diseases, we further validated its adaptability in multimodal multi-disease classification using the Retina Image Bank dataset.[20] Despite the complexity of the dataset, which includes 14 diverse modalities and 84 retinal conditions, including rare diseases, and the significant imbalance in disease conditions, EyeFound is capable of learning, with AUROC =0.566, outperforming RETFound AUROC = 0.530 in this multimodality task (Table 2).

**Performance of Systemic Diseases Diagnosis**

We tested EyeFound's adaptability in systemic disease prediction in four tasks to predict stroke, dementia, PD, and MI in the UK Biobank dataset using ocular images. Our model outperformed RETFound in predicting stroke, dementia, and PD using CFP, achieving AUROC scores of 0.634, 0.547, and 0.571, respectively (all P＜0.05). The same trend is also observed in AUPR, with scores of 0.624, 0.583, and 0.601, respectively (all P＜0.05). Results are provided in Table 3

**Performance of Visual Question Answering**

We fine-tuned EyeFound and RETFound on the VQA task using the Retina Image Bank and evaluated their performance on the OphthalVQA dataset without further training. Detailed results are presented in Table 4. EyeFound outperformed RETFound significantly across all evaluation metrics, including exact matching score, F1 score, Precision, and Recall, as well as the language generation metric BLEU.[25]

**Efficiency of Foundation models**

We evaluated the efficiency of EyeFound on downstream tasks by determining the number of fine-tuning epochs required to achieve the best AUROC on the validation dataset for each task. Supplementary Figure 2 shows the comparison of EyeFound with RETFound and self-supervised imagenet. EyeFound generally converges faster than RETFound and self-supervised ImageNet, requiring fewer epochs to achieve the best performance.

**Model visualization and explanation**

Figure 2 presents the visualization results of EyeFound across different modalities. EyeFound learns to reconstruct the masking regions despite a large masking ratio of 0.8.

**Discussion**

This study introduces EyeFound, a multimodal self-supervised learning-based foundation model that has been pretrained on 11 ophthalmic image modalities using a generative pretraining approach. When fine-tuned for various disease detection tasks, EyeFound has demonstrated significant performance enhancements in diagnosing ocular diseases, as well as systemic and diverse multimodal conditions. By bridging the gap left by single-modality foundational models, EyeFound showcases the potential of harnessing multidimensional data without relying on extensive high-quality labels.

The field of ophthalmology benefits from a wealth of multimodal images, each providing unique and complementary insights. From the anterior to the posterior segments of the eye, different imaging modalities capture various facets of ocular health. [26] External eye photo and Slit lamp imaging captures and enhances the view of the anterior segment of the eye. Corneal topography maps the cornea's shape, while a specular microscope provides magnified visualization of corneal endothelial cells. While CFP records retinal and choroidal changes. FFA and ICGA visualize the dynamic circulation within the retina and choroid, as well as lesions in the retina and retinal pigment epithelium (RPE). FAF indicates the metabolic activity of RPE cells. Ultrasound imaging delves deep into the eye, exposing alterations in the retina, choroid, muscles, and orbital bones. OCT provides sectional views of the retina, highlighting changes in the photoreceptor and RPE layers. RetCam is a specialized form of CFP for infants. Prior works have shown the benefits of integrating different modalities for enhanced vessel segmentation and disease detection.[8-10,27,28] However, there has been a scarcity of foundation models that learn a joint representation from multiple modalities. The earlier model, RETFound, necessitated separate training for CFP and OCT, a limitation that EyeFound overcomes by consolidating multiple modalities into a unified model.

EyeFound's ability to learn multimodal representations is evidenced by its superior performance in multimodal downstream tasks compared to RETFound. In unimodal tasks, EyeFound surpasses RETFound in Glaucoma Fundus detection, while showing comparable results in other diseases. This suggests that joint multimodal training does not compromise the model's general performance when compared with modality-specific training. For tasks that require the combination of multiple modalities, EyeFound consistently surpasses RETFound. The model's effectiveness could be explained by visualizing reconstructed masked regions, which reveals that EyeFound captures the intrinsic relationships among image pixels.

EyeFound utilizes the MAE technique, which enables the model to learn data distributions by predicting masked regions, an idea originally inspired by masked language learning in natural language processing. The convergence of language and vision is an emerging trend for the interactive interpretation of ophthalmic images,[29-31] and our findings intriguingly support this; when the pretrained vision encoder from EyeFound is combined with the large language model (LLM) Llama 2,[23] it excels in zero-shot open-set VQA tasks without any fine-tuning. This opens avenues for future research into the seamless integration of pretrained vision encoder and LLM to achieve efficient specialized VQA.

Despite these promising outcomes, the study has limitations. Firstly, EyeFound was trained on a large-scale dataset from China, and incorporating data from other regions could enhance its diversity and representativeness. Although the model adapts well to multi-country datasets in downstream tasks, its ability to capture all variations in ocular diseases and conditions may be limited, potentially affecting its generalizability to rare or unseen conditions. Secondly, the current model focuses solely on image-based learning, not fully exploiting text labels or descriptions during pretraining, which could mean missing out on valuable information. Future research could investigate methods that combine self-supervised and supervised pretraining, such as contrastive language-image pretraining, to maximize the use of all available data. Lastly, while EyeFound has made strides in integrating multiple modalities into a single model, determining the optimal way to combine and interpret information from different modalities remains an open question for further exploration.

In conclusion, we present EyeFound, an innovative joint multimodal foundation model for ophthalmic imaging. EyeFound has the potential to transform medical AI by learning shared representations for different modalities, enabling efficient adaptation across a broad spectrum of applications. EyeFound excels in a variety of multimodal downstream tasks and shows promise in handling complex open-set zero-shot VQA challenges. Future efforts will aim to further improve the model's performance and explore its clinical application potential.

**Conflicts of interests**

The authors declare no competing interest.

**Funding**

The study was supported by the Global STEM Professorship Scheme (P0046113) and . The sponsor or funding organization had no role in the design or conduct of this research.

**Acknowledgment**

We thank the American Society of Retina Specialists for providing the valuable Retina Image Bank.

**Data availability**

The authors do not have the right to redistribute the dataset used for training EyeFound. The downstream datasets can be accessed by referring to the original paper.

**Code availability**

MAE: https://github.com/facebookresearch/mae, https://github.com/rmaphoh/RETFound_MAE

Llama2: https://github.com/facebookresearch/llama\

**Figure 1 | Overview of EyeFound.** OCT=Optical coherence tomography, CFP=Color fundus photography, FFA=Fundus fluorescein angiography, ICGA=Indocyanine green angiography, FAF=Fundus autofluorescence, DR=Diabetic retinopathy, VQA=Visual question answering, SLO=Scanning laser ophthalmoscopy, FPP=Fundus photography.

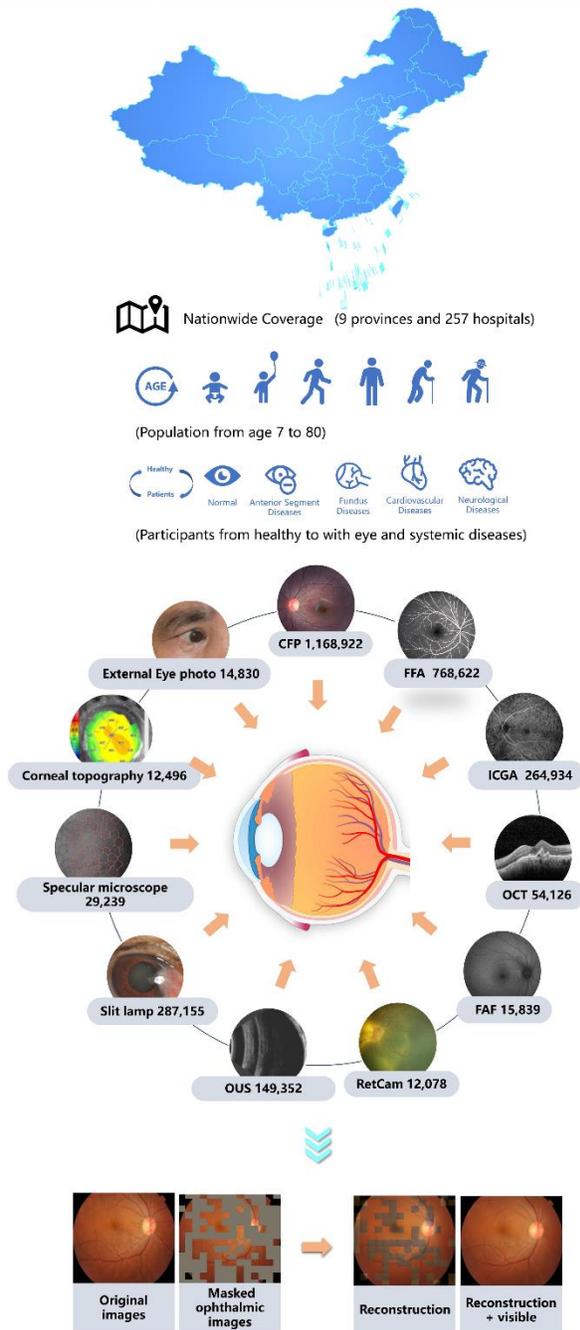
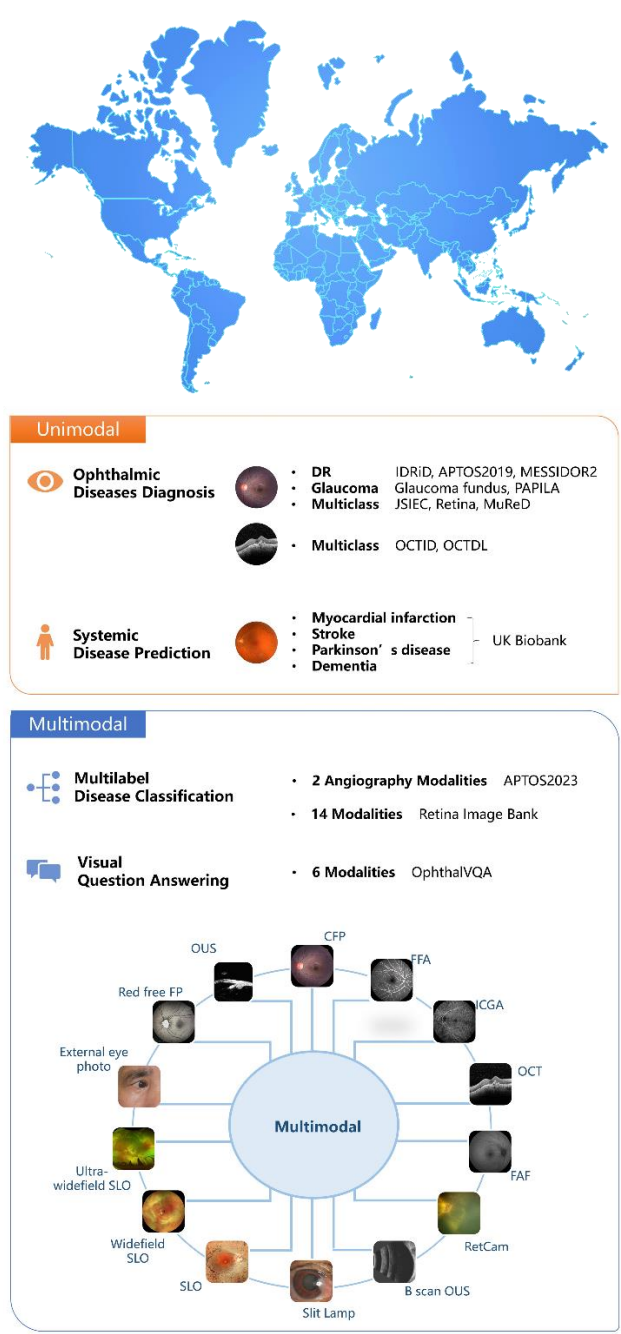

**Figure 2 | Reconstruction of multimodality images of EyeFound**. a, Reconstructed colour fundus photographs from highly masked images. Although with few patches visible, EyeFound and RETFound can infer the retina-specific anatomical structures and disease lesions

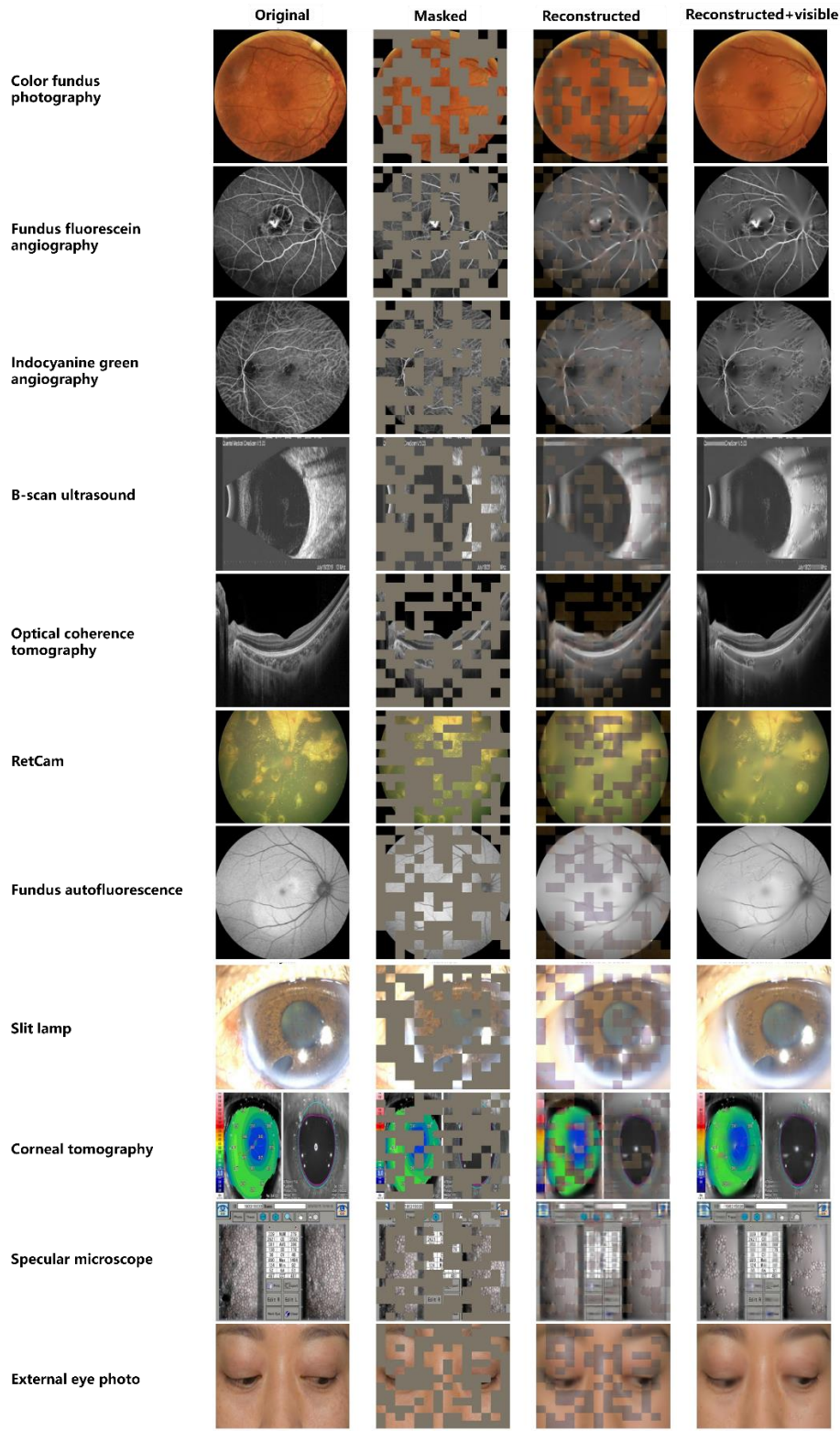

**Table 1 | Dataset for developing EyeFound downstream validation.**

| Dataset | Modality | N |
|---|---|---|
| **Developing EyeFound** | | **2,777,593** |
| **Multi-China** | CFP | 1168922 (42.1%) |
| | FFA | 768622 (27.7%) |
| | Slit lamp | 287155 (10.3%) |
| | ICGA | 264934 (9.5%) |
| | OUS | 149352 (5.4%) |
| | OCT | 54126 (1.9%) |
| | Specular microscope | 29239 (1.1%) |
| | FAF | 15839 (0.6%) |
| | External eye photo | 14830 (0.5%) |
| | Corneal topography | 12496 (0.4%) |
| | RetCam | 12078 (0.4%) |
| **Downstream validation** | | |
| **IDRiD** | CFP | 516 |
| **APTOS2019** | CFP | 3662 |
| **MESSIDOR2** | CFP | 1744 |
| **PAPILA** | CFP | 488 |
| **Glaucoma Fundus** | CFP | 1544 |
| **JSIEC** | CFP | 1000 |
| **Retina** | CFP | 601 |
| **MuReD** | CFP | 2208 |
| **OCTID** | OCT | 572 |
| **OCTDL** | OCT | 2064 |
| **UK Biobank** | CFP | 1436 |
| **AngioReport** | FFA, ICGA | 10520 |
| **Retina Image Bank** | CFP, FFA, ICGA, OCT, FAF, RetCam, B scan OUS, Slit Lamp, SLO, Widefield SLO, Ultra-widefield SLO, External eye, Red free, OUS | 3293 |
| **OphthalVQA** | Slit Lamp, SLO, CFP, OCT, FFA, OUS | 60 |

OCT=Optical coherence tomography, CFP=Color fundus photography, FFA=Fundus fluorescein angiography, ICGA=Indocyanine green angiography, FAF=Fundus autofluorescence, SLO=Scanning laser ophthalmoscopy, OUS=Ocular ultrasound.

**Table 2 | Performance in downstream ocular disease tasks.**

| Dataset | Sample Size | Model | AUROC | 95% CI | AUPR | 95% CI | P value* |
|---|---|---|---|---|---|---|---|
| **Unimodal** | | | | | | | |
| **CFP-DR** | | | | | | | |
| **IDRiD** | 516 | Ours | 0.820 | [0.810, 0.829] | **0.506** | **[0.486, 0.525]** | 0.875 |
| | | RetFound | 0.822 | [0.815, 0.829] | 0.496 | [0.481, 0.511] | |
| **APTOS 2019** | 3662 | Ours | **0.948** | **[0.941, 0.954]** | 0.724 | [0.700, 0.748] | 0.827 |
| | | RetFound | 0.943 | [0.941, 0.944] | 0.726 | [0.721, 0.730] | |
| **MESSIDOR2** | 1744 | Ours | 0.872 | [0.859, 0.884] | 0.652 | [0.632, 0.672] | 0.363 |
| | | RetFound | 0.884 | [0.880, 0.887] | 0.669 | [0.656, 0.683] | |
| **CFP-Glaucoma** | | | | | | | |
| **Glaucoma_Fundus** | 1544 | Ours | **0.955** | **[0.941, 0.969]** | 0.887 | **[0.854, 0.920]** | **0.002** |
| | | RetFound | 0.943 | [0.941, 0.945] | 0.863 | [0.860, 0.867] | |
| **PAPILA** | 488 | Ours | 0.813 | [0.773, 0.853] | 0.663 | [0.616, 0.710] | 0.014 |
| | | RetFound | 0.820 | [0.788, 0.854] | 0.678 | [0.646, 0.709] | |
| **CFP-Multiclass** | | | | | | | |
| **JSIEC** | 1000 | Ours | **0.994** | **[0.992, 0.997]** | 0.917 | **[0.883, 0.952]** | 0.618 |
| | | RetFound | 0.990 | [0.989, 0.991] | 0.884 | [0.878, 0.889] | |
| **Retina** | 601 | Ours | **0.868** | [0.838, 0.897] | **0.743** | **[0.696, 0.789]** | 0.595 |
| | | RetFound | 0.847 | [0.841, 0.853] | 0.697 | [0.691, 0.703] | |
| **MuReD** | 2208 | Ours | **0.875** | **[0.862, 0.881]** | 0.506 | **[0.492, 0.510]** | **<0.001** |
| | | RetFound | 0.861 | [0.855, 0.875] | 0.493 | [0.489, 0.505] | |
| **OCT-Multiclass** | | | | | | | |
| **OCTID** | 572 | Ours | 0.989 | [0.986, 0.992] | 0.954 | [0.938, 0.970] | 0.538 |
| | | RetFound | 0.993 | [0.987, 0.999] | 0.980 | [0.967, 0.993] | |
| **OCTDL** | 2064 | Ours | **0.995** | **[0.989, 0.997]** | 0.943 | **[0.938, 0.962]** | 0.012 |
| | | RetFound | 0.982 | [0.972, 0.992] | 0.903 | [0.862, 0.925] | |
| **Multimodal** | | | | | | | |
| **FFA, ICGA-Multiclass** | | | | | | | |
| **AngioReport** | 10520 | Ours | **0.707** | **[0.684, 0.730]** | 0.297 | **[0.275, 0.319]** | **<0.001** |
| | | RetFound | 0.693 | [0.671, 0.714] | 0.295 | [0.276, 0.314] | |
| **Multimodal-Multiclass** | | | | | | | |
| **Retina Image Bank** | 3293 | Ours | **0.566** | [0.491, 0.641] | **0.165** | [0.100, 0.230] | **<0.001** |
| | | RetFound | 0.530 | [0.484, 0.576] | 0.062 | [0.048, 0.076] | |

AUROC=Area under the receiver operator characteristic curve, AUPR=Area under the precision recall curve, CI=Confidence interval, CFP=Color fundus photography, DR=Diabetic retinopathy, OCT=Optical coherence tomography, FFA=Fundus fluorescein angiography, ICGA=Indocyanine green angiography. Bold highlights where EyeFound outperforms RETFound. *The P-values were calculated using pairwise comparisons of classification results between EyeFound and RETFound.

**Table 3 | Performance in downstream chronic disease tasks.**

| Disease | Model | AUROC | 95% CI | AUPR | 95% CI | P value* |
|---|---|---|---|---|---|---|
| **Stroke** | Ours | **0.634** | **[0.611, 0.657]** | **0.624** | **[0.603, 0.645]** | **<0.001** |
| | RetFound | 0.622 | [0.597, 0.648] | 0.621 | [0.593, 0.648] | |
| **Dementia** | Ours | **0.547** | **[0.491, 0.603]** | **0.583** | **[0.535, 0.631]** | **<0.001** |
| | RetFound | 0.509 | [0.455, 0.564] | 0.554 | [0.520, 0.588] | |
| **PD** | Ours | **0.571** | **[0.503, 0.638]** | **0.601** | **[0.545, 0.658]** | **<0.001** |
| | RetFound | 0.549 | [0.432, 0.665] | 0.593 | [0.513, 0.673] | |
| **MI** | Ours | 0.584 | [0.566, 0.602] | 0.573 | [0.563, 0.583] | 0.002 |
| | RetFound | 0.589 | [0.558, 0.620] | 0.581 | [0.559, 0.603] | |

PD=Parkinson's disease, MI=Myocardial infarction, AUROC=Area under the receiver operator characteristic curve, AUPR=Area under the precision recall curve, CI=Confidence interval. Bold highlights where EyeFound outperforms RETFound. *The P-values were calculated using pairwise comparisons of classification results between EyeFound and RETFound.

**Table 4 | Performance in downstream zero-shot VQA tasks.**

| Model | Exact match score | F1 score | Precision | Recall | BLEU1 | BLEU2 | BLEU3 | BLEU4 | P value* |
|---|---|---|---|---|---|---|---|---|---|
| **Ours** | **0.242** | **0.188** | **0.253** | **0.167** | **0.123** | **0.031** | **0.012** | **0.004** | **<0.001** |
| **RetFound** | 0.221 | 0.157 | 0.229 | 0.132 | 0.094 | 0.015 | 0.003 | 0.001 | |

BLEU=Bilingual evaluation understudy. *The P-values were calculated using pairwise comparisons of classification results between EyeFound and RETFound.